\pdfoutput=1

\documentclass[11pt]{article}

\usepackage[preprint]{acl}

\usepackage{times}
\usepackage{latexsym}
\usepackage{booktabs}  
\usepackage[T1]{fontenc}  
\usepackage[utf8]{inputenc} 
\usepackage{microtype} 
\usepackage{inconsolata} 
\usepackage{graphicx}
\usepackage{enumitem} 
\usepackage{graphicx}
\usepackage{multirow}
\usepackage{amsmath}

\title{\raisebox{-0.45em}{\includegraphics[height=1.5em]{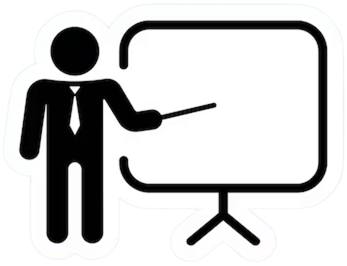}}~PresentAgent: Multimodal Agent for Presentation Video Generation}

\author{
  \textbf{Jingwei Shi$^{1*}$~~
  Zeyu Zhang$^{1*\dag}$~~
  Biao Wu$^{2*}$~~
  Yanjie Liang$^{1*}$}\\
  \textbf{
  Meng Fang$^{3}$~~
  Ling Chen$^{2}$~~
  Yang Zhao$^{4\ddag}$} \\
  $^1$AI Geeks, Australia \\
  $^2$Australian Artificial Intelligence Institute, Australia \\
  $^3$University of Liverpool, United Kingdom\\
  $^4$La Trobe University, Australia\\
  \scriptsize$^{*}$Equal contribution. $^{\dag}$Project lead. $^{\ddag}$Corresponding author: y.zhao2@latrobe.edu.au.
}

\begin{document}

\maketitle
\begin{abstract}
We present PresentAgent, a multimodal agent that transforms long-form documents into narrated presentation videos. While existing approaches are limited to generating static slides or text summaries, our method advances beyond these limitations by producing fully synchronized visual and spoken content that closely mimics human-style presentations. To achieve this integration, PresentAgent employs a modular pipeline that systematically segments the input document, plans and renders slide-style visual frames, generates contextual spoken narration with large language models and Text-to-Speech models, and seamlessly composes the final video with precise audio-visual alignment. Given the complexity of evaluating such multimodal outputs, we introduce PresentEval, a unified assessment framework powered by Vision-Language Models that comprehensively scores videos across three critical dimensions: content fidelity, visual clarity, and audience comprehension through prompt-based evaluation. Our experimental validation on a curated dataset of 30 document–presentation pairs demonstrates that PresentAgent approaches human-level quality across all evaluation metrics. These results highlight the significant potential of controllable multimodal agents in transforming static textual materials into dynamic, effective, and accessible presentation formats.
Code will be available at \url{https://github.com/AIGeeksGroup/PresentAgent}.
\end{abstract}

\section{Introduction}

Presentations are a widely used and effective medium for conveying complex ideas. By combining visual elements, structured narration, and spoken explanations, they enable information to unfold progressively and be more easily understood by diverse audiences~\cite{fu2022doc2ppt}. Despite their proven effectiveness, creating high-quality presentation videos from long-form 
\begin{figure}[ht] 
  \centering
  \includegraphics[width=1\columnwidth]{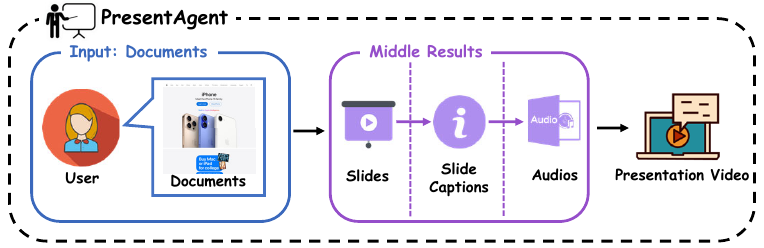} 
  \caption{
  \textbf{Overview of PresentAgent.} It takes documents (e.g., web pages) as input and follows a generation pipeline: (1) document processing, (2) structured slide generation, (3) synchronized caption creation, and (4) audio synthesis. The final output is a presentation video combining visual slides with aligned narration. The purple-highlighted middle results emphasize the system's key transitional outputs during generation.
  }
  \label{fig:easy_workflow}
\end{figure}
documents—such as business reports, technical manuals, policy briefs, or academic papers—typically requires considerable manual effort~\cite{li2023videogen}. This process involves identifying key content, designing slide layouts, writing scripts, recording narration, and aligning all elements into a coherent multimodal output.

Although recent advancements in AI have enabled progress in related areas such as document-to-slide generation ~\cite{fu2022doc2ppt,pptagent,pang2025paper2poster,zhang2024motion} and text-to-video synthesis~\cite{yang2024cogvideox,li2023videogen,xue2025phyt2v,khachatryan2023text2video,he2023animate,solanki2024script,liu2025fpsattention}, a critical gap remains: these methods either produce static visual summaries or generic video clips without structured narration, limiting their effectiveness for structured communication tasks like presentations.

\begin{figure*}[ht]
  \centering
  \includegraphics[width=1\linewidth]{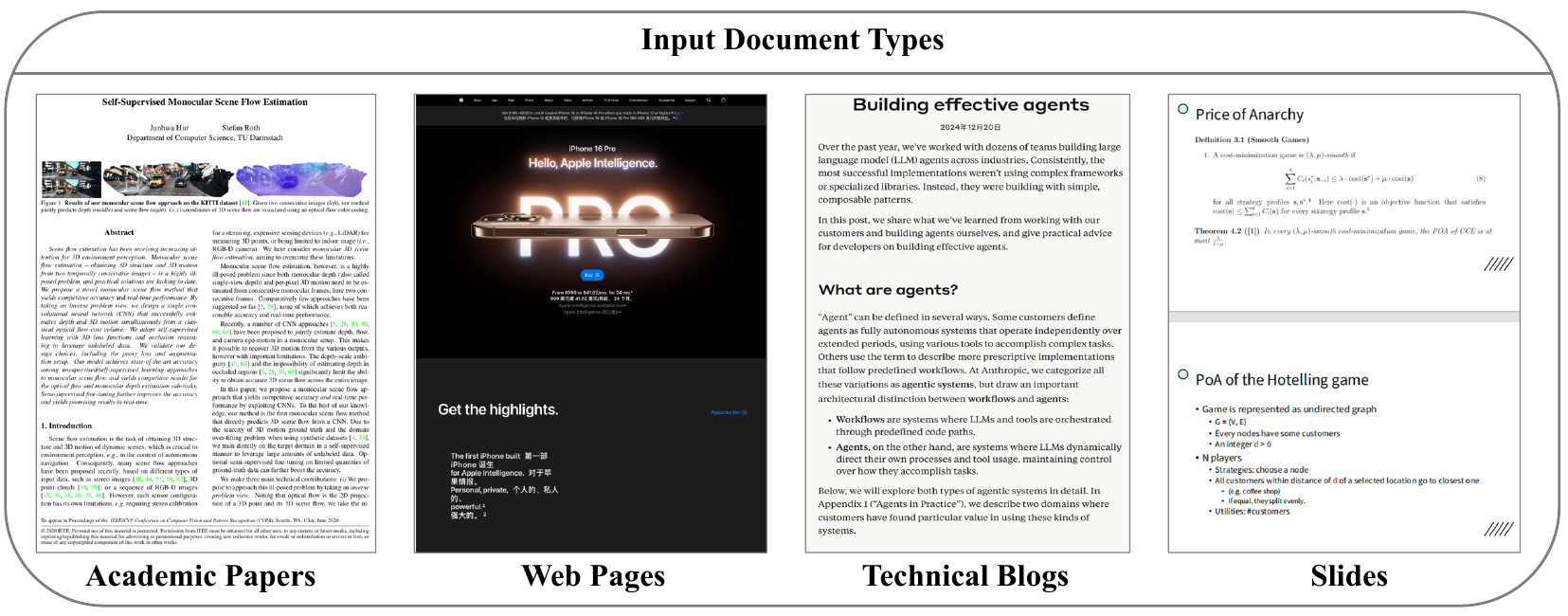}
  \caption{
  Document Diversity in Our Evaluation Benchmark.}
  \label{fig:benchmark}
\end{figure*}

To bridge this gap, we introduce the task of Document-to-Presentation Video Generation, which aims to automatically convert a structured or unstructured document into a narrated video presentation composed of synchronized slides and speech. This task presents unique challenges as it goes beyond  traditional summarization~\cite{lewis2019bart,beltagy2020longformer,chen2021structure,wang2024segmented} or text-to-speech~\cite{tachibana2018efficiently,ren2019fastspeech,popov2021grad,ni2022unsupervised} pipelines by requiring selective content abstraction, layout-aware planning~\cite{wang2025infinity}, and precise multimodal alignment~\cite{li2024llava} between visuals and narration. In contrast to prior work that focuses on either static slide and image generation~\cite{pptagent,deng2025emerging,xie2024show} or audio summarization in isolation, our objective is to produce a fully integrated, viewer-ready video experience that closely mimics how human presenters deliver information in real-world scenarios.

\begin{figure*}[ht]
  \centering
  \includegraphics[width=0.98\linewidth]{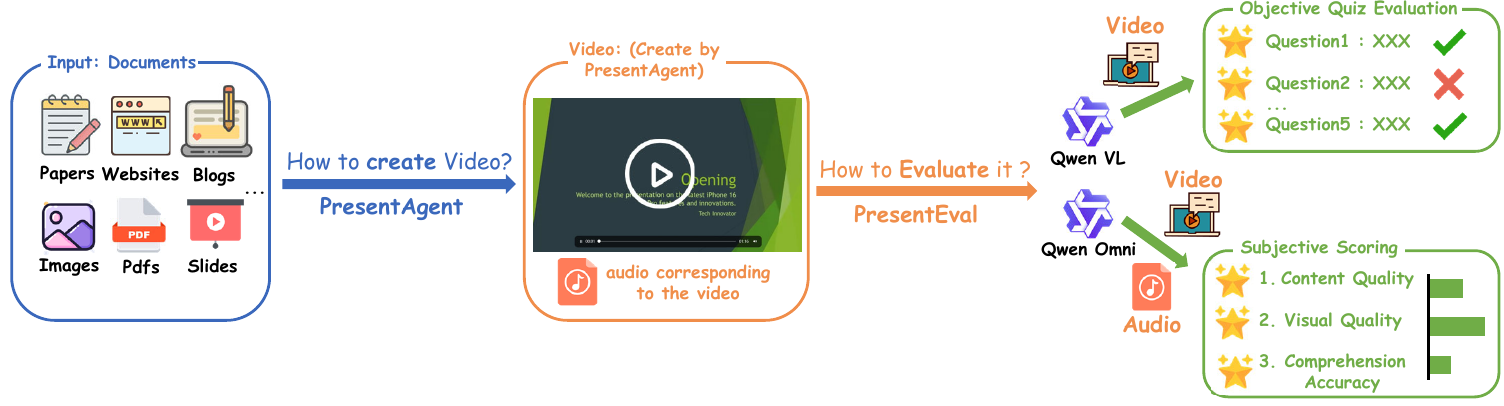}
  \caption{
  \textbf{Overview of our framework.}
   Our approach addresses the full pipeline of document-to-presentation video generation and evaluation. Left: Given diverse input documents—including papers, websites, blogs, slides, and PDFs—PresentAgent generates narrated presentation videos by producing synchronized slide decks with audio. Right: To evaluate these videos, we introduce PresentEval, a two-part evaluation framework: (1) Objective Quiz Evaluation (top), which measures factual comprehension using Qwen-VL; and (2) Subjective Scoring (bottom), which uses vision-language models to rate content quality, visual design, and audio comprehension across predefined dimensions.
  }
  \label{fig:presenteval}
\end{figure*}

To tackle these challenges, we propose a modular generation framework named PresentAgent, which is shown in Figure \ref{fig:easy_workflow}. Given an input document, the system first segments it into semantic blocks through outline planning, then generates layout-guided slide visuals for each block and rewrites the key message into oral-style narration. Subsequently, these are then synthesized into audio and combined with the slide visuals to produce a time-aligned presentation video. Importantly, our pipeline is designed to be domain-adaptable and controllable, enabling broad applicability across document types and presentation styles.

Recognizing the need for rigorous evaluation of such complex multimodal outputs, we curate a test set of 30 human-authored document-video pairs spanning diverse domains, including education, finance, policy, and scientific communication. To comprehensively assess system performance, we further introduce a two-path evaluation strategy that combines fact-based comprehension assessment (via fixed multiple-choice quizzes) and preference-based scoring using vision-language models. This dual-pronged approach captures both objective correctness and subjective quality in video delivery.

Experiment results demonstrate that our method produces fluent, well-structured, and informative presentation videos, approaching human-level performance in both content delivery and viewer comprehension. These findings highlight the potential of combining language models, layout generation, and multimodal synthesis for creating explainable and scalable presentation systems from raw documents.

In general, our contributions are summarized as follows:
\begin{itemize}[leftmargin=*,itemsep=2pt]
    \item We formulate and address the novel task of document-to-presentation video generation, which aims to produce narrated, slide-structured videos from long-form documents across diverse domains.

    \item We propose PresentAgent, a modular generation framework that integrates document parsing, layout-aware slide composition, narration planning, and audio-visual synchronization, enabling controllable and interpretable generation.

    \item We introduce PresentEval, a multi-dimensional evaluation framework powered by Vision-Language Models (VLMs), which scores videos along content, visual, and comprehension dimensions via prompt-based judging.

    \item We create a test set of 30 real-world document–presentation pairs and demonstrate through experiments and ablations that PresentAgent approaches human-level performance and significantly outperforms competitive variants.
\end{itemize}

\section{Related Work}

\subsection{Document-to-Multimodal Generation}

Recent advances in large language models (LLMs) and multimodal generation have sparked growing interest in converting documents into diverse output formats, such as slides, posters, or audio summaries~\cite{xu2025qwen2,wang2025infinity,pang2025paper2poster,sun2024genesis}. Systems like PPTAgent~\cite{zheng2025pptagent} and Doc2PPT~\cite{fu2022doc2ppt} treat document-to-slide generation as a structured summarization problem, focusing on layout-aware slide construction. Other works, such as Paper2Poster~\cite{pang2025paper2poster} extend this idea by producing single-page visual summaries using layout planning and visual feedback. However, these systems typically generate static outputs and do not model time-dependent delivery such as narration or slide progression. Our work builds upon these foundations, but further introduces temporal planning and audio-visual synchronization, enabling the generation of fully narrated presentation videos.

\subsection{Vision-Language Agents}

Recent advances have highlighted the expanding capabilities of vision language models (VLMs) beyond traditional language understanding. Techniques such as ReAct~\cite{react,mmreat,yue2024dots} have shown that LLMs can operate as autonomous agents, capable of step-by-step reasoning and dynamic interaction through code execution~\cite{opendevin,sweagent,yang2024if}, API function calls~\cite{toolformer,octotools,yang2023gpt4tools}, user interface manipulation~\cite{showui,uitars,uivision,wu2024foundations}, and motion generation~\cite{zhang2024motionavatar,zhang2024motion,zhang2024infinimotion,zhang2024kmm,zhang2025motion}. Despite these developments, general-purpose agents still struggle with professional tasks that demand accuracy, domain-specific knowledge, and reliable interaction~\cite{videogui}. A closely related area is slide automation~\cite{autopresent,pptagent}, which agents translate short text prompts into executable Python code to render presentation slides.  In contrast, our proposed presentation video generation task is significantly more challenging: instead of taking a short prompt as input, the system processes an entire long-form document—such as a research paper, product manual, or technical report—and produces a well-structured presentation video with oral-style narration. This task imposes higher demands on content understanding, multimodal alignment, speech generation, and video synthesis. To address these challenges, we design a generation pipeline along with an automatic evaluation framework to systematically assess the generated videos in terms of information delivery, visual quality, and overall comprehensibility.

\section{Presentation Benchmark}

The benchmark supports evaluation not only of fluency and fidelity, but also of downstream comprehension. Following the methodology introduced in Paper2Poster~\cite{pang2025paper2poster}, we construct a quiz-style evaluation protocol in (\S\ref{sec:presenteval}), where vision-language models are asked to answer factual content questions using only the generated video (slides + narration), simulating an audience’s understanding. Human-authored videos are used as reference standards for both score calibration and upper-bound comparison. As shown in Figure \ref{fig:benchmark}, our benchmark encompasses four representative document types (academic papers, web pages, technical blogs, and slides) paired with human-authored videos, covering diverse real-world domains like education, research, and business reports.
\begin{table*}[t]
\small
\centering
\resizebox{\linewidth}{!}{
\begin{tabular}{cl}
\toprule
Prensentation of Web Pages & What is the main feature highlighted in the iPhone's promotional webpage? \\
\midrule 
A. & A more powerful chip for faster performance \\
B. & A brighter and more vibrant display \\
C. & An upgraded camera system with better lenses \\
D. & A longer-lasting and more efficient battery \\

\midrule

Prensentation of Academic Paper & What primary research gap did the authors aim to address by introducing the FineGym dataset? \\
\midrule 
A. & Lack of low-resolution sports footage for compression studies \\
B. & Need for fine-grained action understanding that goes beyond coarse categories \\
C. & Absence of synthetic data to replace human annotations \\
D. & Shortage of benchmarks for background context recognition \\

\bottomrule
\end{tabular}
}
\caption{Prompt of evaluation via Objective Quiz Evaluation. Each question set is manually created based on the actual document content, with a focus on topic recognition, structural understanding, and key argument identification. These questions evaluate how well the generated video communicates the source material.}
\label{tab:videoquiz_questions}
\end{table*}

\begin{table*}[ht]
\centering
\resizebox{\linewidth}{!}{
\begin{tabular}{ll}
\toprule
\textbf{Video} & \textbf{Scoring Prompt} \\
\midrule
Narr. Coh. &
\textit{“How coherent is the narration across the video? Are the ideas logically connected and easy to follow?”} \\
\midrule
Visual Appeal &
\textit{“How would you rate the visual design of the slides in terms of layout, aesthetics, and overall quality?”} \\
\midrule
Comp. Diff. &
\textit{“How easy is it to understand the presentation as a viewer? Were there any confusing or contradictory parts?”} \\
\midrule
\textbf{Audio} & \textbf{Scoring Prompt} \\
\midrule
Narr. Coh. &
\textit{“How coherent is the narration throughout the audio? Are the ideas logically structured and easy to follow?”} \\
\midrule
Audio Appeal &
\textit{“How pleasant and engaging is the narrator’s voice in terms of tone, pacing, and delivery?”} \\
\midrule
Comp. Diff. &
\textit{“How easy is it to understand the spoken content? Were there any unclear or confusing parts in the audio?”} \\
\bottomrule
\end{tabular}
}
\caption{Prompt of evaluation via Subjective Scoring. Each prompt targets a specific dimension—narrative coherence, visual/audio appeal, or comprehension difficulty—and is designed to guide vision-language models in assessing presentations from a human-centric perspective. Abbreviations: Narr. Coh. = Narrative Coherence; Comp. Diff. = Comprehension Difficulty.}
\label{tab:audio_quality_prompts}
\end{table*}

We adopt a unified, model-based evaluation framework to assess the generated presentation videos. All evaluations are conducted using a vision-language model, guided by dimension-specific prompts tailored to different assessment objectives. The framework consists of two complementary components: (1) objective quiz evaluation, which measures factual accuracy through multiple-choice question answering; and (2) subjective scoring, which rates Content Quality, Visual or Audio Quality, and Comprehension Clarity on a 1–5 scale. Together, these metrics provide a comprehensive assessment of both the quality and informativeness of the generated videos.

\subsection{Doc2Present Dataset}

To support the evaluation of document to presentation video generation, we curate the Doc2Present Benchmark, a diverse dataset of document–presentation video pairs spanning multiple domains. Unlike prior benchmarks focused on research abstracts or slide generation, our dataset includes documents such as business reports, product manuals, policy briefs, and instructional texts, each paired with a human-crafted presentation video.

\paragraph{Data Source.} We collect 30 high-quality video samples from public platforms, educational repositories, and professional presentation archives. Each video follows a structured narration format, combining slide-based visuals with synchronized voiceover. We manually align each video with its source document and ensure the following conditions are met: (1) the content structure of the video follows that of the document; (2) the visuals convey document information in a compact, structured form; and (3) the narration and slides are well-aligned temporally.

\paragraph{Data Statistics.} The average document length is 3,000–8,000 words, while the corresponding videos range from 1 to 2 minutes and contain 5-10 slides. This setting highlights the core challenge of the task: transforming dense, domain-specific documents into effective and digestible multimodal presentations.

\subsection{PresentEval}
\label{sec:presenteval}

To assess the quality of generated presentation videos, we adopt two complementary evaluation strategies: Objective Quiz Evaluation and Subjective Scoring, as shown in Figure \ref{fig:presenteval}. For each video, we provide the vision-language model with the complete set of slide images and the full narration transcript as a unified input—simulating how a real viewer would experience the presentation. In Objective Quiz Evaluation, the model answers a fixed set of factual questions to determine whether the video accurately conveys the key information from the source content. In Subjective Scoring, the model evaluates the video along three dimensions: the coherence of the narration, the clarity and design of the visuals, and the overall ease of understanding. All evaluations are conducted without ground-truth references and rely entirely on the model’s interpretation of the presented content.

\paragraph{Objective Quiz Evaluation}

To evaluate whether a generated presentation video effectively conveys the core content of its source document, we use a fixed-question comprehension evaluation protocol. Specifically, we manually design five multiple-choice questions for each document, tailored to its content. These questions focus on key aspects such as topic recognition, structural understanding, and main argument extraction. As shown in Table \ref{tab:videoquiz_questions}, during evaluation, a vision-language model is given the video, including both visual frames and audio transcript, and asked to answer the five questions. Each question has four options, with one correct answer, annotated based on a human-created reference video. The final comprehension score (ranging from 0 to 5) reflects how many questions the model answered correctly, serving as a direct measure of how well the video communicates the original document.

\paragraph{Subjective Scoring}

To evaluate the quality of generated presentation videos, we adopt a prompt-based assessment using vision-language models. Instead of relying on human references or fixed metrics, we ask the model to evaluate each video from a viewer’s perspective, using its own reasoning and preferences. The evaluation focuses on three aspects: coherence of narration, clarity and aesthetics of visuals, and overall ease of understanding. The model is shown the video and audio, and gives a score (1–5) with a brief explanation for each aspect. This enables scalable, consistent, and human-aligned evaluation without manual references. As shown in Table \ref{tab:audio_quality_prompts}, we design different prompts for different modalities and tasks to ensure targeted and effective assessment.

\begin{figure*}[ht]
  \centering
  \includegraphics[width=0.98\linewidth]{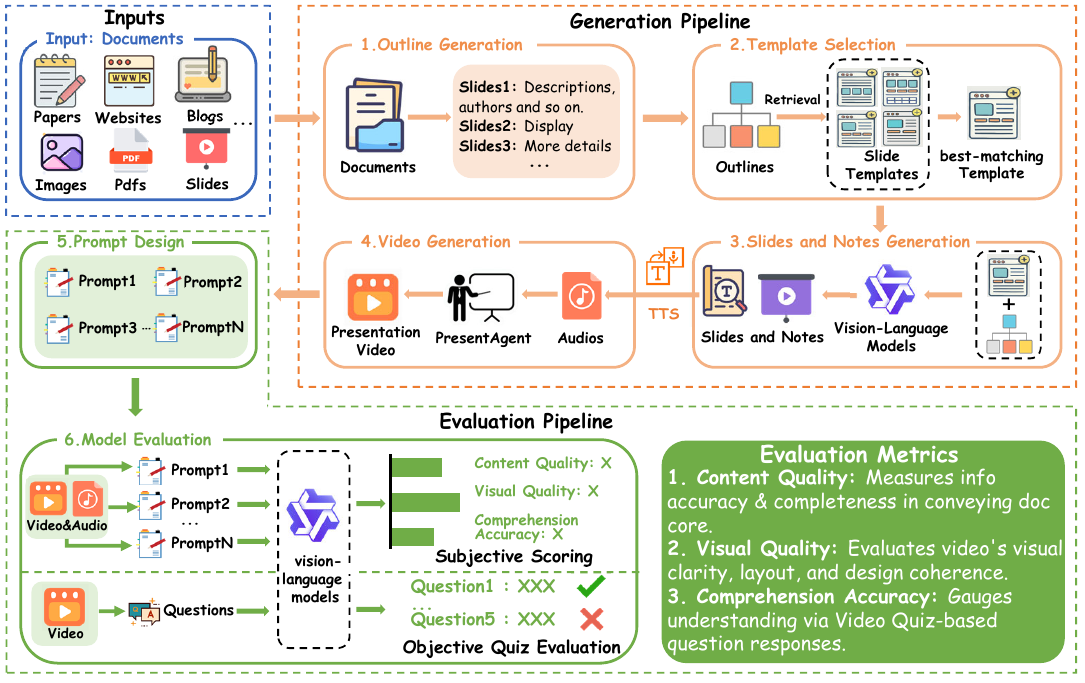}
  \caption{
    \textbf{Overview of the PresentAgent framework.} 
    Our system takes diverse documents (e.g., papers, websites, PDFs) as input and follows a modular generation pipeline. It first performs outline generation (Step 1) and retrieves the most suitable template (Step 2), then generates slides and narration notes via a vision-language model (Step 3). The notes are converted into audio via TTS and composed into a presentation video (Step 4). To evaluate video quality, we design multiple prompts (Step 5) and feed them into a VLM-based scoring pipeline (Step 6) that outputs dimension-specific metrics.
    }
  \label{fig:overview}
\end{figure*}

\section{PresentAgent} 

To convert a long-form document into a narrated presentation video, we design a multi-stage generation framework that mirrors how human presenters prepare slides and talk tracks, as shown in Figure \ref{fig:overview}. Our method proceeds in four steps: segmenting the document into semantic units, composing slides with layout-aware structures, generating oral-style narration for each slide and assembling the visual and audio components into a synchronized video. This modular design supports controllability, interpretability, and multimodal alignment, enabling both high-quality generation and fine-grained evaluation. The following sections describe each component in detail.

\subsection{Problem Formulation}

Our method is designed to transform a long-form document into a structured presentation video through a multi-stage generation pipeline. We provide a formal description to highlight the key difference between our approach and conventional slide-based methods.

Conventional approaches often focus on generating slide elements \( S \) directly from a document chunk \( C \), as in Equation~\ref{eq:conventional}, where each element includes text or image content, layout attributes, and visual style:
\[
S = \{e_1, e_2, ..., e_n\} = f(C) \tag{1} \label{eq:conventional}
\]

In contrast, we treat the entire document \( D \) as a globally structured input and generate a presentation in three steps: 
(1) a sequence of semantic segments \( \{C_1, ..., C_K\} \) via outline planning, 
(2) a set of slides \( \{S_1, ..., S_K\} \), each paired with a narrated audio track \( T_k \) generated by first producing a slide-specific script and then converting it to speech, and (3) a video \( V \) composed of visual and audio content aligned over time.  This is defined as:

\[
V = \texttt{Compose}(\{(S_1, T_1), ..., (S_K, T_K)\}) = g(D) \tag{2} \label{eq:ours}
\]

Rather than editing predefined templates or layouts, our system first identifies high-level structure in the document and then generates slide visuals and narration from scratch. This pipeline supports controllability, modular evaluation, and multimodal alignment for downstream comprehension and quality assessment.

\begin{table*}[t]
\centering
\resizebox{\linewidth}{!}{
\begin{tabular}{lc|c|cccc|cccc}
\toprule
\multirow{2}{*}{\textbf{Method}} & \multirow{2}{*}{\textbf{Model}} 
& \multirow{2}{*}{\textbf{Quiz Accuracy }} 
& \multicolumn{4}{c|}{\textbf{Video Score}} 
& \multicolumn{4}{c}{\textbf{Audio Score}} \\
\cmidrule(r){4-7} \cmidrule(r){8-11}
& & & Content & Visual & Comp. & Mean & Content & Audio & Comp. & Mean \\
\midrule
Human & Human & 0.56 & 4.0 & 4.6 & 4.8 & 4.47 & 4.8 & 4.6 & 5.0 & 4.80 \\

\midrule

PresentAgent & Claude-3.7-sonnet & 0.64 & 4.0 & 4.0 & 4.0 & 4.00 & 4.2 & 4.6 & 4.8 & 4.53 \\
PresentAgent & Qwen-VL-Max & 0.52 & 4.2 & 4.8 & 4.4 & 4.47 & 4.6 & 4.2 & 5.0 & 4.60 \\
PresentAgent & Gemini-2.5-pro & 0.52 & 4.2 & 4.4 & 4.4 & 4.33 & 4.2 & 4.0 & 4.8 & 4.33 \\
PresentAgent & Gemini-2.5-flash & 0.52 & 4.2 & 5.0 & 3.8 & 4.33 & 4.2 &4.2 & 4.8 & 4.40 \\
PresentAgent & GPT-4o-Mini & 0.64 & 4.8 & 4.6 & 4.6 & 4.67 & 4.0 & 4.4 & 4.8 & 4.40 \\
PresentAgent & GPT-4o & 0.56 & 4.0 & 4.2 & 3.6 & 3.93 & 4.2 & 4.4 & 4.8 & 4.47 \\

\bottomrule
\end{tabular}
}
\caption{
Detailed evaluation results on the 5-document test set. Fact-based evaluation includes accuracy on five fixed quiz questions (Q1–Q5). Preference-based evaluation includes 1–5 scale scores for content fidelity, visual design, and overall clarity. Each Quality Score group has a calculated mean column.
}
\label{tab:detailed_eval}
\end{table*}

\subsection{Slide Planning and Composition}

Our slide generation module is inspired by the editing-based paradigm proposed in PPTAgent~\cite{zheng2025pptagent}, which formulates presentation construction as a structured editing process over HTML-like layouts. While PPTAgent focuses on producing editable \texttt{.pptx} slides, our goal is to generate visually coherent, narration-ready slide frames for downstream video synthesis. We re-implement the core idea in a self-contained pipeline tailored to multimodal synchronization.

We begin by segmenting the input document into coherent content blocks using a lightweight LLM-based parser. Each block is assigned a corresponding slide type such as bullet slide, figure-description, or title-intro, and matched with a predefined layout schema encoded in HTML. Unlike retrieval-based template matching, our system uses semantic and structural cues to map content to layout patterns in a rule-guided manner.

To populate the slide, we define a set of editable operations such as \texttt{replace\_text}, \texttt{insert\_image}, and \texttt{add\_list}, which are applied to the layout structure. These instructions are generated by prompting a language model with the content block and layout constraints. Slides are then rendered into static visual frames using \texttt{python-pptx} or HTML-based renderers.

\subsection{Narration and Audio Synthesis}

To transform the static slides into an engaging presentation, we generate a spoken narration for each slide and synthesize it into audio. The process involves two components: narration script generation and text-to-speech synthesis.

For each content block corresponding to a slide, we prompt a language model to generate a concise, oral-style narration. The model is instructed to rewrite the key message of the slide into natural spoken language, avoiding dense text or technical jargon. We apply length control to ensure each narration falls within a target duration, typically between 30 and 150 seconds. Once the narration script is obtained, we synthesize the corresponding audio using a text-to-speech system. Each narration audio is paired with its slide and timestamped, forming the basis for synchronized video rendering in the next stage.

\subsection{Video Assembly}

In the final stage, we assemble the slide images and narration audio into a coherent, time-aligned presentation video. Each slide frame is displayed for the duration of its corresponding audio segment, with optional transitions between segments. We use video processing libraries such as \texttt{ffmpeg} to compose the visual and audio tracks. Each slide is rendered as a static frame, and the narration is added as synchronized voiceover audio. The output is a fully rendered video file in standard formats such as \texttt{.mp4}, suitable for presentation, sharing, or further editing. This stage completes the transformation from a raw document into a narrated, structured presentation video.

\section{Experiments}

We conduct experiments to evaluate the effectiveness of our proposed system in generating high-quality, narrated presentation videos. Given the novelty of the task, our focus is not on competing with existing baselines, but rather on assessing the performance of our full system relative to human-created presentations.
Comprehension accuracy is determined based on performance in the PresentEval task.

\subsection{Evaluation Setup}

We construct a test set consisting of 30 long-form documents, each paired with a manually created presentation video that serves as a human-level reference. These documents span a diverse range of topics, including education, product explanation, research overviews, and policy briefings. For each document, we generate a corresponding presentation video using our full generation pipeline.

All videos, both human-created and machine-generated, are evaluated using our unified evaluation framework, PresentEval. Each synthesized video is approximately two minutes in length. However, due to the current lack of a single multimodal model capable of jointly assessing visual and audio quality for videos longer than two minutes, we adopt a split evaluation strategy.

In the Objective Quiz stage, we use Qwen-VL-2.5-3B~\cite{wang2024qwen2} to evaluate the accuracy of the entire video using a fixed set of multiple-choice comprehension questions. In the Subjective Scoring stage, we extract short video/audio segments and evaluate them individually to assess quality in a more focused and scalable manner, using Qwen-Omni-7B~\cite{xu2025qwen2}.

Both models are guided by dimension-specific prompts and score each video or audio sample along three axes: Content Quality, Visual Quality, and Comprehension Accuracy. 

\begin{figure*}[ht]
  \centering
  \includegraphics[width=1\linewidth]{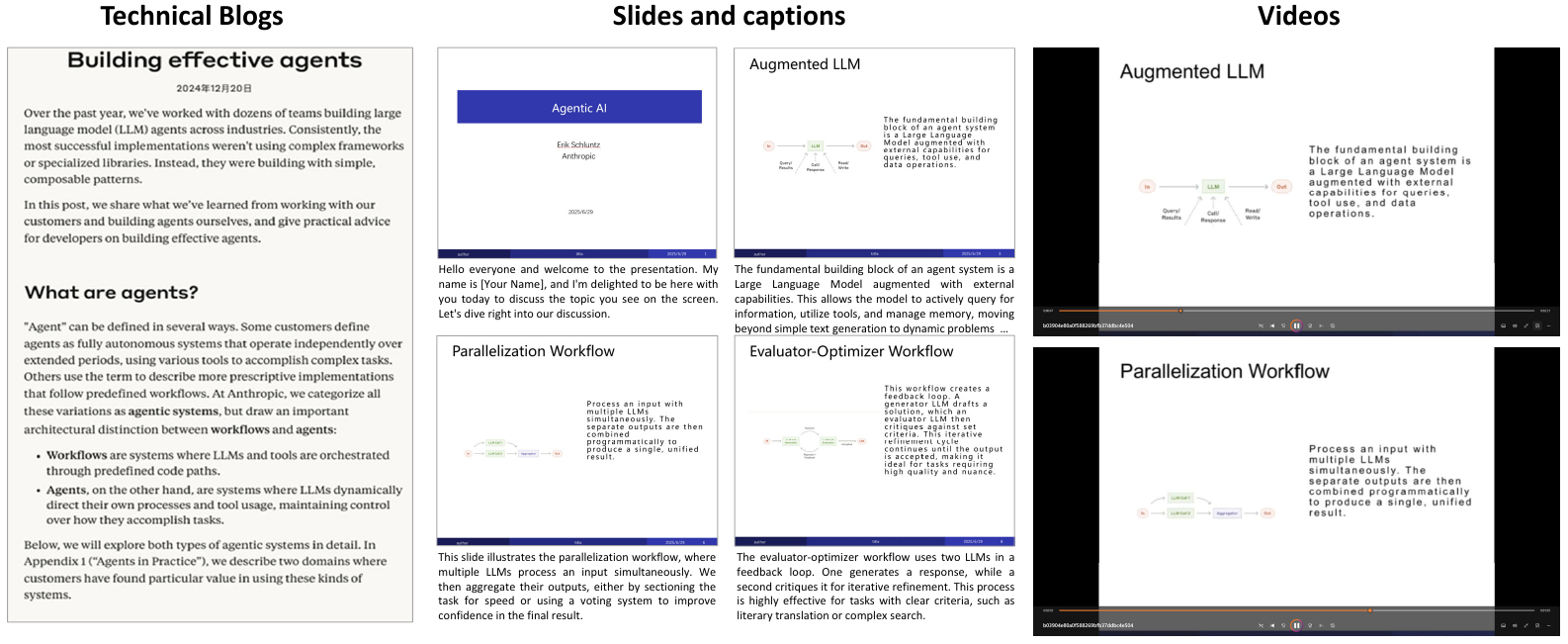}
  \caption{
    \textbf{PresentAgent Demo.} 
     Automatically generates academic-style slides and narrated videos from research papers, streamlining the transformation from written content to engaging visual presentations.
    }
  \label{fig:demo}
\end{figure*}

\subsection{Implementation Details}

PresentAgent adopts a highly modular multimodal-generation architecture. At the language-understanding and generation layer, we run six primary LLM back ends in parallel—GPT-4o, GPT-4o-mini, Qwen-VL-Max, Gemini-2.5-Flash, Gemini-2.5-Pro, and Claude-3.7-Sonnet—and select or ensemble them on-the-fly with a dynamic routing policy that weighs input length, conversational complexity, and latency budget. For visual-language evaluation, we introduce the lightweight VLM Qwen-VL-2.5-3B-Instruct to score slide layout, chart readability, and cross-modal consistency, feeding its self-critique back into generation. Speech synthesis is unified on MegaTTS3, which outputs 24 kHz, 16-bit high-fidelity narration and supports prosody-tag controls for fine-grained rate, pitch, and emotion adjustment.

The experimental pipeline converts any input document—PDF, Markdown, DOCX, or web snapshot through three automated stages:

\begin{enumerate}[leftmargin=*,itemsep=2pt]

\item Structured parsing \& re-ordering that maps content to a hierarchical topic–subtopic tree.

\item Per-slide generation with the chosen LLM, producing a PowerPoint deck containing titles, bullet points, graphic placeholders, and Alt-Text, while retrieving and inserting relevant images for key nouns.

\item Synchronized narration generation with MegaTTS3 in Chinese or English, followed by an FFmpeg script that assembles a 1080 p video with fade-in/out transitions and optional captions.

\end{enumerate}

\subsection{Main Results}

Table \ref{tab:detailed_eval} presents evaluation results, covering both factual comprehension (Quiz Accuracy) and preference-based quality scores for video and audio outputs. In terms of quiz accuracy, most PresentAgent variants perform comparably to or better than the human reference (0.56), with Claude-3.7-sonnet~\cite{claude3} achieving the highest accuracy at 0.64, suggesting strong alignment between the generated content and the source document. Other models such as Qwen-VL-Max~\cite{bai2025qwen2} and Gemini-2.5-flash~\cite{gemini2.5flash} scored slightly lower (0.52), indicating room for improvement in factual grounding.

In terms of subjective quality, human-created presentations still lead with the highest video and audio scores overall. However, several PresentAgent variants show competitive performance. For example, GPT-4o-Mini~\cite{achiam2023gpt} achieves top scores in video content and visual appeal (both at or near 4.8), while Claude-3.7-sonnet~\cite{claude3} delivers the most balanced audio quality (mean 4.53). Interestingly, Gemini-2.5-flash~\cite{gemini2.5flash} scores highest in visual quality (5.0) but lower in comprehension, reflecting a trade-off between aesthetics and clarity. These results highlight the effectiveness of our modular pipeline and the usefulness of our unified PresentEval framework in capturing diverse aspects of presentation quality.

\subsection{Analysis}

Figure \ref{fig:demo} Presents a full example of a PresentAgent-auto-generated presentation video, showing a technical blog turned into a narrated presentation. The system identifies structural segments (e.g., introduction, technical explanations) and generates slides with oral-style captions and synchronized speech, covering topics like “parallelization workflow” and “agent system architecture” to demonstrate its ability to keep technical accuracy while delivering content clearly and conversationally.

The result highlights PresentAgent’s capacity to extract key semantics, generate structured visuals, and produce expressive TTS narration, with the coherent, time-aligned output validating its effectiveness and versatility in technical communication and product demos.

\section{Discussion}

In this work, we synthesized presentation-style videos that integrate visual slides, textual narration, and spoken audio, simulating realistic multimodal communication scenarios. While our current evaluation focuses on the individual quality of each modality—such as visual clarity, textual relevance, and audio intelligibility—these dimensions are treated independently. However, in real-world applications, the effectiveness of communication often hinges on the semantic and temporal coherence across modalities.

Future research should thus move beyond isolated assessments and aim toward fusion-aware understanding and evaluation. This entails not only modeling the interactions and alignment among image, audio, and text modalities, but also enabling the system to reason over their combined meaning. Existing models like ImageBind offer a unified embedding space for multiple modalities, but lack the capacity for high-level inference and semantic comprehension.

A promising direction lies in bridging representation alignment with multimodal reasoning, by integrating aligned modality encoders with powerful language models. This would allow the system to jointly perceive, interpret, and respond to complex multimodal inputs—such as explaining a visual concept based on both audio narration and visual cues, or identifying inconsistencies across modalities. Developing such reasoning-capable, fusion-aware models will be critical for advancing robust, coherent multimodal understanding in real-world applications.

\section{Limitation and Future Work}
Our work faces two key constraints: (1) Due to the high computational costs of commercial LLM/VLM APIs (e.g., GPT-4o and Gemini-2.5-Pro), evaluation was limited to five academic papers, potentially underrepresenting the document diversity shown in our benchmark (Figure~\ref{fig:benchmark}); (2) PresentAgent currently generates static slides without dynamic animations/effects due to architectural constraints in video synthesis and trade-offs between generation speed and visual quality, as noted in ChronoMagic-Bench's temporal coherence studies. Future improvements could involve lightweight distillation models and physics-aware rendering engines.
Our future work will focus on three directions: first, making extensive attempts by leveraging a larger number of open-source models as the foundational bases—spanning various architectural designs, capability scopes, fine-tuning strategies and incorporating a much broader range of document categories for both generation and evaluation processes, covering diverse domains, formats, real-world and practical application scenarios to ensure thorough, comprehensive and in-depth exploration of the capability of our proposed PresentAgent; second, integrating dynamic animation capabilities by optimizing video synthesis architectures, balancing speed-quality trade-offs, and testing with complex scene transitions; third, exploring lightweight distillation and physics-aware rendering to enhance generation efficiency, realism, and adaptability to varied hardware environments.

\section{Conclusion}

We presented PresentAgent, a modular system for the novel task of transforming long-form documents into narrated presentation videos. By systematically addressing slide planning, narration synthesis, and synchronized rendering, PresentAgent enables controllable and reusable multimodal outputs across diverse document types. To support rigorous evaluation, we introduced a benchmark of document–video pairs and proposed dual assessment strategies: factual quizzes and preference-based vision-language scoring. Experimental results, including extensive ablations and model comparisons, show that PresentAgent produces coherent, engaging, and highly informative presentations that closely approach human quality. Our work demonstrates the promise of integrating language and vision models for generating explainable and viewer-friendly content from complex, varied texts, establishing a cornerstone for future research in automated and controllable multimodal content generation for education, business and accessibility.

\end{document}